\documentclass[11pt]{article}

\usepackage[final]{acl}

\usepackage{times}
\usepackage{latexsym}

\usepackage[T1]{fontenc}

\usepackage[utf8]{inputenc}

\usepackage{microtype}

\usepackage{inconsolata}

\usepackage{graphicx}

\usepackage[shortlabels,inline]{enumitem}
\usepackage{tikz,lipsum}
\usepackage[most]{tcolorbox}
\usepackage{verbatim}
\usepackage{amssymb}
\usepackage{minted}
\usepackage{fancyvrb}
\usepackage{ragged2e}
\usepackage[dvipsnames]{xcolor}
\usepackage{amsmath}
\usepackage{booktabs}
\usepackage{tabularray}
\usepackage{arydshln}
\usepackage{stmaryrd}
\usepackage{marvosym}
\usepackage{makecell}
\usepackage{multirow}
\usepackage{longtable}
\usepackage{adjustbox}
\usepackage{dsfont}
\usepackage{xspace}
\usepackage{multicol}
\usepackage{multirow}
\usepackage{colortbl}

\usepackage{cleveref}
\crefname{section}{\S}{\S}
\crefname{table}{Table}{Tables}
\crefname{figure}{Fig.}{Figs.}
\crefname{algorithm}{Alg.}{}
\crefname{ALC@unique}{Line}{Lines}
\crefname{equation}{Eq.}{Eqs.}
\crefname{appendix}{App.}{Apps.}
\crefformat{section}{\S#2#1#3} 
\usepackage{soul}


\usepackage{todonotes}

\NewDocumentCommand{\prompt}{O{} m +m}{%
  \begin{promptbox}[#1]{#2}{}%
  #3
  \end{promptbox}
}

\newtcolorbox[auto counter,number within=section]{promptbox}[2][]{
    coltitle=white,
    colframe=black,
    colback=black!5!white,
    enhanced jigsaw,
    breakable,
    pad at break*=2mm,
    fontupper=\footnotesize,
    fontlower=\footnotesize,
    fonttitle=\small,
    left=4pt,
    right=4pt,
    title={Prompt \thetcbcounter: #2},
    label={#1},
}

\usepackage{lineno}
\usepackage{amsmath}
\usepackage{amssymb}
\usepackage{booktabs}
\usepackage{pifont}
\usepackage{color}
\usepackage[table]{xcolor}
\usepackage{bm}
\usepackage{multirow}
\usepackage{soul}
\usepackage{bbm}
\usepackage{xurl}
\usepackage{graphicx}
\usepackage[shortlabels,inline]{enumitem}
\usepackage{tikz,lipsum}
\usepackage{ragged2e}
\usepackage[dvipsnames]{xcolor}
\usepackage{tabularray}
\usepackage{arydshln}
\usepackage{stmaryrd}
\usepackage{marvosym}
\usepackage{makecell}
\usepackage{multirow}
\usepackage{nameref}
\usepackage{amsfonts}
\usepackage{dsfont}
\usepackage[T1]{fontenc}

\usepackage{soul}
\definecolor{acllow}{HTML}{B22222}
\definecolor{aclhigh}{HTML}{1B7A3E}
\newcommand{\errhl}[1]{{\sethlcolor{red!18}\hl{#1}}}
\newcommand{\fixhl}[1]{{\sethlcolor{green!20}\hl{#1}}}

\usepackage{xspace}
\newcommand{\ours}{\textsc{\tt VC-Inspector}\xspace}

\title{\ours: Advancing Reference-free Evaluation \\of Video Captions with Factual Analysis}

\author{
    Shubhashis Roy Dipta\hspace{.1em}$^{\,1}$\thanks{These authors contributed equally to this work.}\quad
    Tz-Ying Wu\hspace{.1em}$^{\,2\ast}$\quad
    Subarna Tripathi\hspace{.1em}$^{\,2}$ \vspace{1.5mm}\\
    $^1$\textbf{University of Maryland, Baltimore County} \;
    $^2$\textbf{Intel}\vspace{1.5mm}\\
    {\texttt{sroydip1@umbc.edu}\quad \texttt{tz-ying.wu@intel.com}}
}

\begin{document}
\maketitle

\begin{abstract}
We propose \ours, a lightweight, open-source large multimodal model (LMM) for \textit{reference-free} evaluation of video captions, with a focus on factual accuracy. Unlike existing metrics that suffer from limited context handling, weak factuality assessment, or reliance on proprietary services, \ours offers a reproducible and fact-aware alternative that aligns closely with human judgments.
To enable robust training and interpretable evaluation, we introduce a systematic framework for generating captions with controllable factual errors, paired with graded quality scores and explanatory annotations.
Experiments demonstrate that \ours achieves state-of-the-art correlation with human judgments, generalizing across diverse domains (e.g., VATEX-Eval, Flickr8K-Expert, and Flickr8K-CF benchmarks) and revealing the potential for caption improvement.
Project page: \url{https://dipta007.github.io/VC-Inspector}.
\end{abstract}

\section{Introduction}
Video captions summarize salient objects and actions in videos, supporting downstream applications such as question answering, event localization, and retrieval~\cite{videoQA2024,Qian2024MomentorAV,gabeur2020multi}. Most evaluation metrics rely on {\it reference-based} protocols~\cite{papineni_bleu_2002,banerjee_meteor_2005,lin_rouge_2004,vedantam_cider_2015,bert-score}, which compare captions to human-written references. These approaches are costly to scale and often fail to capture semantic equivalence.

\begin{figure}[t!]
    \centering
    \includegraphics[width=.95\linewidth]{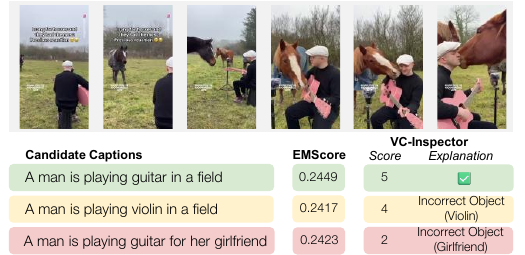} \\
    \vspace{3pt}
    \includegraphics[width=.95\linewidth]{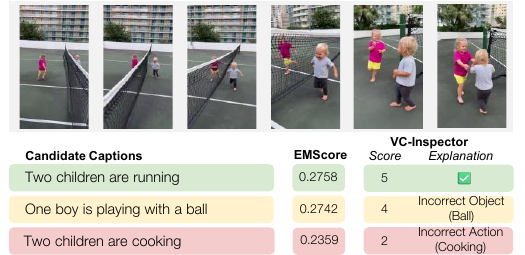}
    \caption{Existing {\it reference-free} metrics like EMScore~\cite{shi_emscore_2022} often fail to detect factual inaccuracies and lack a consistent scoring scale. \ours addresses these limitations by providing {\it factually grounded}, \textbf{interpretable evaluations}.}
    \label{fig:teaser}
\end{figure}

Evaluating captions for in-the-wild videos requires {\it reference-free} protocols that do not rely on ground truth captions, yet this remains underexplored.
Existing metrics~\cite{shi_emscore_2022,sarto2023positive} typically measure visual-language alignment using pretrained multimodal embeddings~\cite{radford_learning_2021}, but
are limited by text encoder context length, and
lack a consistent scoring scale, making interpretation difficult (\cref{fig:teaser}).
Recently, large proprietary models such as GPT-4o are used to rate captions on a fixed scale~\cite{tong_g-veval_2024}, but these approaches depend heavily on prompt engineering and are not fully reproducible. Additionally, most prior methods are image-centric, rendering them suboptimal for video content.

In this work, we aim to develop a {\it reference-free} evaluation metric for video captions that removes dependency on human-annotated references while remaining robust and intuitive for humans.
Our approach is grounded in {\bf factual accuracy}, as factual elements like objects and actions are critical for video understanding.
Ideally, a reliable metric should degrade scores proportionally to factual errors. For example, for a video showing a little girl sitting on a chair, the caption ``A little girl is sleeping on a chair'' should receive a higher score than ``A woman is sleeping on a chair'', because the latter deviates more from the video content. However, existing metrics like EMScore~\cite{shi_emscore_2022} often fail to capture even basic factual inaccuracies, such as incorrect objects or actions. 

To address these limitations, we propose \ours, built on top of a lightweight, open-source large multimodal model (LMM) trained to assess caption quality based on factual correctness. Unlike previous metrics that provide only a score, \ours also generates explanations for its judgments, improving interpretability.
One key challenge is the lack of captions with varying degrees of factual quality for training.
We overcome this by introducing a novel data generation framework powered by a large language model (LLM), which
\textbf{controllably} modifies factual elements in ground-truth captions from ActivityNet-Captions~\cite{krishna_dense-captioning_2017}.
This process yields 
{\tt ActivityNet-FG-It}, a dataset of 44K instances for instruction tuning. ~\cref{fig:main} illustrates our data generation and training framework.

We comprehensively evaluate the quality of \ours across multiple complementary settings.
We first examine its consistency and reliability on two synthetic datasets, {\tt ActivityNet-FG-Eval} and {\tt YouCook2-FG-Eval}, which contain captions of diverse quality generated using the same pipeline as our instruction-tuning data. Results show that \ours produces stable quality estimates across visual domains and caption lengths.
We further assess its correlation with human judgments on two video caption datasets, VATEX-Eval~\cite{shi_emscore_2022} and YouCook2-Eval, and extend the evaluation to image caption benchmarks, Flickr8K-Expert and Flickr8K-CF~\cite{hodosh2013framing}, by viewing images as an extreme case of short videos.
Across all datasets, \ours outperforms prior {\it reference-free} metrics and even surpasses most {\it reference-based} metrics, highlighting its strong generalization capability.
In addition, we evaluate \ours on two hallucination benchmarks, FOIL-COCO~\cite{shekhar2017foil} and ActivityNet-FOIL~\cite{shi_emscore_2022}, where FOIL captions contain object errors. Explicitly trained for object and action grounding, \ours performs strongly on both benchmarks.
Beyond score prediction, we further show that the explanatory outputs of \ours not only enhance quality estimation but also enable caption refinement.  

To summarize, this work makes the following contributions:
\begin{itemize}
    \item We introduce a scalable pipeline for synthesizing video captions with controllable factual errors, enabling large-scale training and evaluation, without costly human annotation.
    \item We propose \ours, a {\it fact-aware}, {\it reference-free} video caption evaluator that jointly predicts quality scores and factual error explanations, enhancing interpretability and guiding caption improvement.
    \item We demonstrate that \ours achieves high correlations with human judgments,
    outperforming existing {\it reference-free} methods and rivaling {\it reference-based} metrics across video and image caption benchmarks, showing strong cross-domain generalization.
\end{itemize}

\section{Related Works}

\paragraph{Text-only metrics based on references.}
Traditional rule-based metrics such as BLEU~\cite{papineni_bleu_2002}, METEOR~\cite{banerjee_meteor_2005}, ROUGE~\cite{lin_rouge_2004}, and CIDEr~\cite{vedantam_cider_2015} primarily capture syntactic similarity and fail to reflect semantic meaning.
SPICE~\cite{anderson_spice_2016,li_factual_2023} introduces semantic parsing of objects, attributes, and relations for both reference and candidate captions, but struggle with paraphrases.
This has motivated the embedding-based methods, such as BERTScore~\cite{bert-score} and its variant~\cite{yi2020improving}, which consider the intrinsic variance between multiple ground truth captions.
More recently, CLAIR~\cite{chan_clair_2023} explores an LLM-as-a-Judge approach for image caption evaluation, demonstrating stronger correlation with human judgments than the above metrics.
However, all these metrics depend on reference captions for evaluation and ignore the visual content.

\paragraph{Image-augmented metrics.}
To overcome the limitation of text-only metrics, image-aware approaches align captions with visual input.
VIFIDEL~\cite{madhyastha2019vifidel} matches object names between images and captions, but is restricted to discrete categories and cannot capture motion dynamics in videos.
Visual-language model (VLM) based metrics offer richer semantic alignment through continuous representations.
ViLBERTScore~\cite{lee_vilbertscore_2020} extends BERTScore using ViLBERT~\cite{lu2019vilbert}, but still requires comparison to reference captions. UMIC~\cite{lee2021umic} and CLIP-based metrics~\cite{hessel_clipscore_2022,jiang_tiger_2019,cui_learning_2018,sarto2023positive, wada_polos_2024,zeng_hicescore_2024,lee_fleur_2024,shi_emscore_2022,liu2023models} relax this constraint by directly modeling image-caption semantic alignment via contrastive learning, though remain limited by the context length of their text encoders.
Recently, \citet{maeda_vision_2024} compare the candidate caption to the VLM-derived visual context with a LLM.
However, these methods focus on static images and do not model the temporal dynamics, which limits their effectiveness for video caption evaluation.

In summary, text-based methods require costly, high-quality reference captions, while image-based extensions remain less effective for video content. These challenges underscore the need for {\it reference-free} video caption evaluation tailored to video inputs, a topic that is relatively less studied compared to text-based and image-caption evaluation.  
EMScore~\cite{shi_emscore_2022} was an early attempt that supports evaluating video captions without a reference. Although it considers both frame-level and video-level embeddings, they are derived from an image-based encoder~\cite{radford_learning_2021}, and its text encoder is constrained by short context length. PAC-S~\cite{sarto2023positive} and FactVC~\cite{liu2023models} augment EMScore with positive and negative data synthesis, respectively. While sharing some similarity to this work, they consider only a single level of corruption, with binary (positive/negative) differentiation, whereas our method incorporates captions with varying degrees of quality, enabling a more nuanced evaluation. 
In addition, these metrics produce a single scalar score without offering explanations of their judgments, posing challenges for interpreting the quality assessment.
More recently, G-VEval~\cite{tong_g-veval_2024} extends G-Eval~\cite{liu_g-eval_2023} to video by stitching together three frames per video.
However, it is unclear whether this image-based prompting generalizes to longer or more dynamic videos. Moreover, the dependence on large proprietary models such as GPT-4o limits its scalability and reproducibility for widespread use. 
In contrast, \ours is explicitly trained for video-level factual grounding, produces interpretable explanations, and avoids reliance on proprietary large models, addressing key limitations of prior work.

\section{Methodology}

\begin{figure}[!t]
    \centering
    \includegraphics[width=\linewidth]{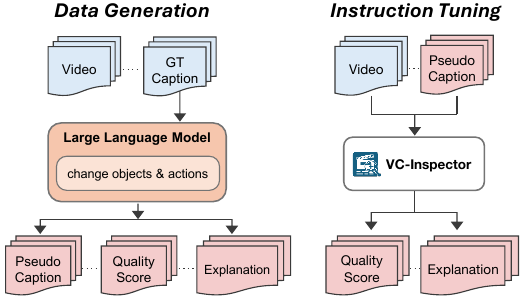}
    \caption{(left) We present a data generation pipeline designed to systematically create synthetic video captions with diverse quality scores, along with explanations for the assigned scores. (right) This dataset was subsequently used for instruction tuning the \ours.}
    \label{fig:main}
\end{figure}

\subsection{Overview}
Video caption quality estimation aims to quantify the correctness of a caption $\hat{X}={\cal M}(V)$ given a video $V$, where ${\cal M}$ denotes a video captioning model.
Prior work falls into two categories: {\it reference-based} and {\it reference-free} metrics. The former compares the candidate caption $\hat{X}$ against the ground truth caption $X$, whereas the latter eliminates this dependency.  

We focus on the {\it reference-free} setting, evaluating $(V,\hat{X})$ without $X$, as assuming reference captions are always available is impractical.  
Existing metrics in this category typically leverage pretrained visual-language embeddings~\cite{radford_learning_2021} to assess the semantic alignment between $V$ and $\hat{X}$. However, these methods are limited by the text encoder’s context length and lack an interpretable scoring scale. 
On the contrary, recent work~\cite{tong_g-veval_2024} employs large proprietary models (e.g., GPT-4o) to support longer contexts, but these approaches rely heavily on prompt engineering and are not fully reproducible.

We argue that a reliable evaluation protocol should be {\bf factually grounded, interpretable, and scalable}. It must reflect the correctness of factual elements (e.g., objects and actions) in the caption relative to the video. For example, captions with missing objects or incorrect actions should receive lower scores, and those with multiple errors should be penalized more severely.
Furthermore, evaluation should be intuitive for humans and reproducible for broad adoption.

To address these limitations, we propose \ours, which leverages a lightweight, open-source large multimodal model (LMM) as the backbone for long-context reasoning and generalized video feature extraction. Our hypothesis is that the strong visual-language reasoning capability of LMMs can transfer effectively to this task.
However, pretrained LMMs do not inherently produce factually grounded assessments and therefore require targeted supervision (\cref{tab:result_on_vatexeval}).
A key challenge is the scarcity of annotated captions exhibiting diverse degrees of factual accuracy for effective training.
To overcome this, we introduce a novel data generation pipeline powered by large language models (LLMs) that systematically synthesizes candidate captions paired with graded quality scores and factual error explanations.
In this work, we focus explicitly on object and action inaccuracies, which constitute the predominant sources of factual errors in video captions.
The overall data generation and training framework is illustrated in \cref{fig:main}.

\subsection{Data Generation}\label{sec:data_creation}

\begin{figure}[t!]
    \centering
    \includegraphics[width=.95\linewidth]{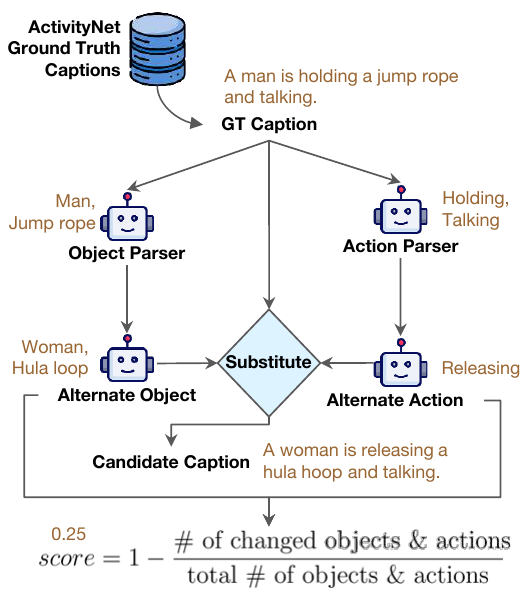}
    \caption{Data generation pipeline to create a synthetic dataset for training \ours. While both ``talking'' and ``holding'' were identified as actions, only ``holding'' was sampled for replacement in the synthetic dataset.}
    \label{fig:data_generation}
    \vspace{-4pt}
\end{figure}

To create captions with controlled factual errors, we employ \texttt{Llama-3.3-70B-Instruct}
to systematically alter objects and actions in ground truth captions from a supervised video caption dataset. Using this approach, we construct \texttt{ActivityNet-FG-It}, an instruction-tuning dataset for factual grounding derived from the ActivityNet-Captions~\cite{krishna_dense-captioning_2017} training set.
The data generation pipeline is illustrated in \cref{fig:data_generation}.  

\paragraph{Caption generation.}
Given a ground truth caption $X$, we first prompt the LLM to extract the set of objects ${\cal O}=\{o_1,...,o_M\}$ and actions ${\cal A}=\{a_1,...,a_N\}$. We then randomly sample $K\sim Unif(0, M)$ objects and $L\sim Unif(0, N)$ actions to replace, forming a subset $\cal{R}\subseteq {\cal O}\cup{\cal A}$, where $|{\cal R}|=K+L$ and $Unif(a,b)$ denotes a discrete uniform distribution over integers $\{a,...,b\}$.  

For each object $o_i\in \cal{R}$, we instructed the LLM to generate an alternative object $\tilde{o}_i$ belonging to the same category (e.g., replacing ``car'' with ``truck'') but with a distinct meaning,
ensuring non-trivial substitutions (e.g., replacing ``car'' with ``building''). Similarly, for each action $a_j\in \cal{R}$, we acquire an alternative action $\tilde{a}_j$ that the subject could perform but conveys a different meaning. For example, changing ``standing'' to ``jumping.'' Note that, the LLM was instructed to generate only the plausible action that the subject can perform. Finally, the selected objects and actions in $\cal{R}$ are then replaced with their corresponding alternatives with the LLM, resulting in the pseudo caption $\tilde{X}$.  

This process also yields a factual error explanation for $\tilde{X}$, enabling fine-grained supervision. The detailed prompts and explanation formats are provided in \cref{app:prompts}.

\paragraph{Scoring.}
After generating a pseudo caption, assigning an intuitive quality score is essential to align with human expectations.
While pretrained embeddings like visual-language model (VLM) can estimate this score, they lack a fixed scale, are constrained by context length, and do not guarantee factual grounding.
Instead, we adopt a \textbf{deterministic} scoring mechanism based on factual accuracy:
\begin{align*}
    score &= 1 - \frac{\text{\# of changed objects \& actions}}{\text{total \# of objects \& actions}} \nonumber\\
    &= 1 - \frac{|{\cal R}|}{|{\cal O}| + |{\cal A}|}.\label{eq:quality}
\end{align*}
Since $\cal{R}\subseteq {\cal O}\cup{\cal A}$ and $|{\cal O}\cup{\cal A}|=|{\cal O}| + |{\cal A}|$,
the score lies on a fixed scale between 0 and 1, ensuring captions with more incorrect elements receive lower scores.
Since LLMs are inherently unreliable when comparing floating-point values \citep{spathis_first_2023}, we discretize the score into a 1-5 range ($score = round(score \times 4 + 1)$), consistent with standard human annotation protocols (e.g., Likert-style ratings in VATEX-Eval).
While this may induce minor loss of information (e.g., $0.87 \rightarrow 4$ and $0.88 \rightarrow 5$),
this conversion preserves ordinal relationships and error severity while maintaining the underlying deterministic structure.

Moreover, \ours is jointly trained to produce textual explanations that explicitly identify incorrect objects and actions, providing fine-grained supervision beyond the scalar score.

\paragraph{Post-processing.}
We repeat the caption generation and scoring process to create 10 pseudo captions per ground truth caption, yielding 374K pseudo captions derived from 37,396 video-caption pairs.
The randomized replacement of objects and actions ensures coverage across the full range of the possible scores, as the number of replacements directly influences the semantic deviation. However, this naturally results in a skewed and non-uniform score distribution. To mitigate the potential bias during training, we apply a balanced sampling strategy, resulting in a refined subset of approximately 218K pseudo captions with uniform representation across five score categories.  

Due to computational constraints, training on the full 218K instances would require multiple weeks. Therefore, we further sample 44K captions (8.8K per label) from the balanced dataset for instruction tuning, which we refer to as {\tt ActivityNet-FG-It}.
This subset preserves category balance and maintains the diversity, while offering a tractable size for experimentation ($\sim32$ GPU hours using A100).  
Importantly, although we use \texttt{Llama} as the generator and perform subsampling as a practical design choice, our data generation pipeline is model-agnostic and applies across model scales and caption datasets, enabling scalable dataset construction at arbitrary sizes.

\begin{figure}[t!]
    \centering
    \includegraphics[width=\linewidth]{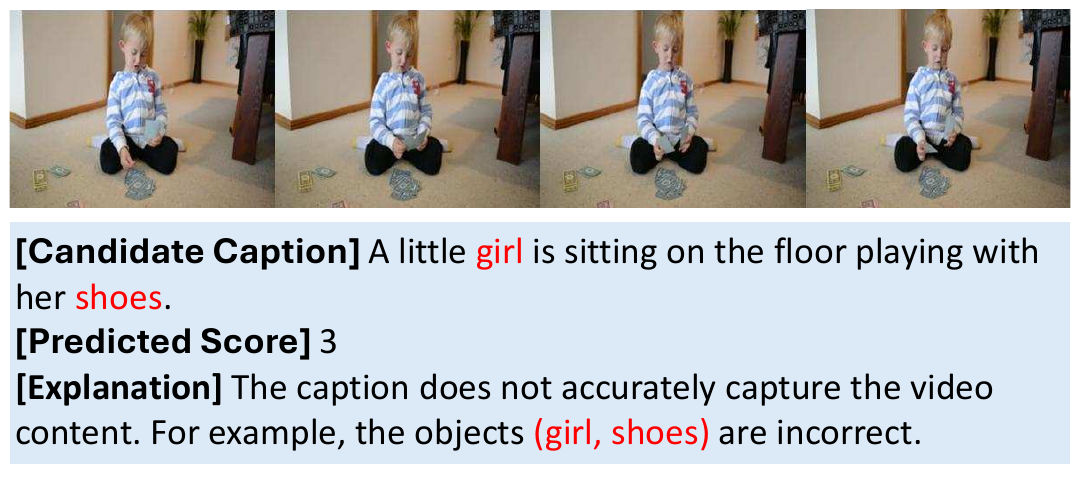} \\
    \vspace{-8pt}
    \captionof{figure}{
    Visual example on VATEX-Eval. \ours produces quality assessments consistent with ground truth scores, and 
    factual error explanations (highlighted in red). More examples are in \cref{app:visual}.}
    \label{fig:qual_example_1}
    \vspace{-8pt}
\end{figure}

\subsection{Training of \ours}
While evaluation metrics are preferred to be lightweight, we employ the 3B/7B version of \texttt{Qwen2.5-VL} as the foundation model by finetuning it with \texttt{ActivityNet-FG-It}.
To preserve the generalization capability of video features, we freeze the pretrained video encoder and the visual-language projector, and only finetune the model parameters in the LLM component with low-rank adaptation~\cite{hu2022lora}.
Given a video-caption pair $(V,\tilde{X})$, the model predicts a quality score $S\in\{1,...,5\}$ along with the corresponding explanation $E$, i.e.,
\begin{align*}
    [S, E] = \ours(V, \tilde{X}).
\end{align*}
The explanation $E$ is formatted in natural language using information collected during data generation (i.e., the list of altered objects and actions) following the template described in \cref{app:prompts}.
This can serve as extra supervision for the model to learn factual grounding and provide interpretable reasoning for this model-based evaluator at test time. \cref{fig:qual_example_1} presents an example of the \ours output.
During training, the model is optimized with the standard language modeling loss~\cite{liu2023llava}.

\section{Experiments} \label{sec:experiments}

\subsection{Experimental Settings}

\paragraph{Training.}
We train \ours with the proposed instruction tuning dataset, \texttt{ActivityNet-FG-It}, which comprises 44K video-caption pairs, along with their quality scores and explanations.

\paragraph{Evaluation.}
The experiments evaluate the consistency and reliability of \ours (Section~\ref{sec:syn_eval}), its correlation to human judgments (Section~\ref{sec:human_corr}), its sensitivity to the targeted factual errors (Section~\ref{sec:foil}), and the contribution of individual components (Section~\ref{sec:ablate_data_syn} and \ref{sec:ablate_exp}).  
Following prior works on evaluation metrics~\cite{shi_emscore_2022,tong_g-veval_2024}, we report Kendall's correlation ($\tau_b$) and Spearman's rank correlation ($\rho$) with ground truth scores in our main experiments.

\paragraph{Implementation details.}
Each video is uniformly sampled into 32 frames and resized to a resolution of $224 \times 224$ for both training and testing. For any image dataset, we treat them as one-frame videos.
\ours is developed for two model sizes, 3B and 7B, initialized from their corresponding {\tt Qwen2.5-VL} pretrained weights. In all experiments, we train the model on 4 NVIDIA-A100 GPUs with a global batch size of 128 and a learning rate of 1e-4.
We set both alpha and rank to 32 for the low-rank adaptation with a dropout rate of 0.05. During inference, we use a temperature of 0.0 for reproducibility. 
The training hyperparameters are provided in \cref{app:train_hyperparams}.

\begin{table}[t!]
    \centering
    \resizebox{\columnwidth}{!}{
    \setlength{\tabcolsep}{4pt}
    \begin{tabular}{@{}l c c c c@{}}
        \toprule
        & \multicolumn{2}{c}{\textbf{\textit{ActivityNet-FG-Eval}}} & \multicolumn{2}{c}{\textbf{\textit{YouCook2-FG-Eval}}}\\
        \cmidrule(lr){2-3} \cmidrule(lr){4-5} 
        Metric & $\tau_b$ & $\rho$ & $\tau_b$ & $\rho$\\
        \midrule
        EMScore~\cite{shi_emscore_2022} & 28.94 & 40.77 & 20.21 & 29.24 \\
        CLIPScore~\cite{hessel_clipscore_2022} & 28.10 & 39.65 & 18.00 & 26.14 \\
        {\tt Qwen2.5-VL}-3B~\cite{bai_qwen25-vl_2025} & 37.91 & 47.80 & 37.16 & 47.17 \\ \midrule
        \ours-3B & {\bf 49.53} & {\bf 62.01} & {\bf 44.29} & {\bf 55.31} \\
        \bottomrule
    \end{tabular}
    }
    \vspace{-5pt}
    \caption{Correlation scores on the synthetic {\tt ActivityNet-FG-Eval}~\cite{caba_heilbron_activitynet_2015} and {\tt YouCook2-FG-Eval}~\cite{ZhXuCoCVPR18} datasets. The best score is \textbf{bolded}.}
    \label{tab:result_anet-yc2}
\end{table}

\subsection{Baselines}\label{sec:baselines}
Baselines are organized into three categories:
{\bf i) Language-based metrics}
solely rely on text reference without considering the visual input. Representative metrics are
BLEU~\cite{papineni_bleu_2002}, ROUGE~\cite{lin_rouge_2004}, METEOR~\cite{banerjee_meteor_2005}, CIDEr~\cite{vedantam_cider_2015}, BERTScore~\cite{bert-score}, SPICE~\cite{anderson_spice_2016}, SPICE-Factual~\cite{li_factual_2023}, Soft-SPICE ~\cite{li_factual_2023} and CLAIR~\cite{chan_clair_2023}.
{\bf ii) Image-augmented metrics} incorporate images as references (alongside reference captions), for better capturing the semantic alignment between the visual input and the candidate caption, e.g., CLIPScore~\cite{hessel_clipscore_2022}, EMScore~\cite{shi_emscore_2022}, PAC-S~\cite{sarto2023positive}, FactVC~\cite{liu2023models}, FLEUR~\cite{lee_fleur_2024}, and G-VEVAL~\cite{tong_g-veval_2024}. They usually support both {\it reference-free} and {\it reference-based} settings.
We report results for both settings for completeness, but our primary focus is on the {\it reference-free} setting, which is more practical for real-world, in-the-wild videos. 
{\bf iii) Video-based metrics} employ a video encoder to incorporate full video sequences. To the best of our knowledge, no existing metric in the literature falls into this category. Therefore, we adapt CLIPScore~\cite{hessel_clipscore_2022} to the recent advent of ViCLIP~\cite{wang_internvid_2024} as a stronger baseline, ViCLIPScore. Additionally, we compare \ours against the base model it builds upon, the vanilla {\tt Qwen2.5-VL} model, to highlight the benefit of fine-tuning for evaluative reasoning.

\begin{table}[!t]
    \centering
    \setlength{\tabcolsep}{2.5pt}
    \resizebox{\columnwidth}{!}{
    \begin{tabular}{@{}lcccccc@{}}
        \toprule
        & \multicolumn{2}{c}{\textbf{\textit{No Reference}}} & \multicolumn{2}{c}{\textbf{\textit{1-Reference}}} & \multicolumn{2}{c}{\textbf{\textit{9-References}}} \\
        \cmidrule(lr){2-3} \cmidrule(lr){4-5} \cmidrule(lr){6-7}
        Metric & $\tau_b$ & $\rho$ & $\tau_b$ & $\rho$ & $\tau_b$ & $\rho$ \\
        \midrule
        \rowcolor{lightgray}\multicolumn{7}{l}{\it Language-based} \\
        BLEU\_1~\cite{papineni_bleu_2002} & - & - & 12.65 & 16.52 & 28.70 & 36.88 \\
        BLEU\_4~\cite{papineni_bleu_2002} & - & - & 12.44 & 14.81 & 22.76 & 25.60 \\
        ROUGE-L~\cite{lin_rouge_2004} & - & - & 12.94 & 16.89 & 23.94 & 31.06 \\
        METEOR~\cite{banerjee_meteor_2005} & - & - & 16.68 & 21.80 & 27.64 & 35.76 \\
        CIDEr~\cite{vedantam_cider_2015} & - & - & 17.62 & 23.02 & 27.92 & 36.18 \\
        BERTScore~\cite{bert-score} & - & - & 15.24 & 19.82 & 25.05 & 32.37 \\
        SPICE~\cite{anderson_spice_2016} & - & - & 14.80 & 18.78 & 27.41 & 35.40 \\
        SPICE-factual~\cite{li_factual_2023} & - & - & 13.59 & 17.04 & 26.05 & 35.58 \\
        Soft-SPICE~\cite{li_factual_2023} & - & - & 21.25 & 27.61 & 36.31 & 46.41 \\
        CLAIR$^*$~\cite{chan_clair_2023} & - & - & 36.00 & - & 34.80 & - \\
        
        \rowcolor{lightgray}\multicolumn{7}{l}{\it Multimodal - image-based} \\
        CLIPScore~\cite{hessel_clipscore_2022} & 22.33 & 29.09 & 27.39 & 35.49 & 35.21 & 45.28 \\        
        EMScore~\cite{shi_emscore_2022} & 22.88 & 29.79 & 28.63 & 37.05 & 36.66 & \textbf{47.00} \\
        FactVC~\cite{liu2023models} & 22.79 & 29.69 & 28.78 & \textbf{37.22} & 36.18 & 46.33 \\
        PAC-S$^*$~\cite{sarto2023positive} & 25.10 & - & 31.40 & - & 38.10 & - \\
        G-VEval$^*$~\cite{tong_g-veval_2024} & 39.40 & - & \textbf{44.90} & - & \textbf{48.10} & - \\
        
        \rowcolor{lightgray}\multicolumn{7}{l}{\it Multimodal - video-based} \\
        ViCLIPScore~\cite{wang_internvid_2024} & 30.92 & 39.86 & - & - & - & - \\      
        {\tt Qwen2.5-VL}-3B~\cite{bai_qwen25-vl_2025} & 31.29 & 36.43 & - & - & - & - \\
        {\tt Qwen2.5-VL}-7B~\cite{bai_qwen25-vl_2025} & 34.70 & 39.40 & - & - & - & - \\
        \midrule
        \ours-3B & 37.99 & 42.45 & - & - & - & - \\
        \ours-7B & \textbf{42.58} & \textbf{45.99} & - & - & - & - \\
        \bottomrule
    \end{tabular}
    }
    \vspace{-5pt}
    \caption{Human correlation scores on the VATEX-Eval~\cite{shi_emscore_2022} dataset. $^*$ indicates results reported from the original paper.
    The best of each column section is \textbf{bolded}.
    Please note that this work focuses on the {\it No Reference} setting.
    \textbf{Our best model outperforms all other models in this setting, while remaining competitive with reference-based metrics.}}
    \label{tab:result_on_vatexeval}
\end{table}

\subsection{Consistency and Reliability of Quality Estimates}\label{sec:syn_eval}
We first examine whether the quality estimation of \ours is consistent across visual domains. Following the same data generation pipeline in Section~\ref{sec:data_creation}, we create two evaluation datasets, \texttt{ActivityNet-FG-Eval} and \texttt{YouCook2-FG-Eval}, according to the ActivityNet~\cite{krishna_dense-captioning_2017} test set and YouCook2~\cite{ZhXuCoCVPR18} validation set, respectively. Both datasets contain pseudo captions with varying degrees of factual errors.

\cref{tab:result_anet-yc2} presents the correlation between the metric scores and ground truth annotations in these two datasets.
\ours, specifically finetuned for factual grounding, consistently outperforms baselines in differentiating incorrect captions. Notably, although it is only trained on {\tt ActivityNet-FG-It}, its quality estimation generalizes effectively across visual domains. This suggests that the proposed data generation pipeline produces stable and reliable quality estimates, rather than noisy outputs.

\begin{table}[t!]
    \centering
    \resizebox{.8\columnwidth}{!}{
    \setlength{\tabcolsep}{9pt}
    \begin{tabular}{@{}l c c@{}}
    \toprule
    Metric & $\tau_b$ & $\rho$ \\ 
    \midrule
    PAC-S (ViT-B)~\cite{sarto2023positive} & 27.8 & 34.3 \\
    PAC-S (ViT-L)~\cite{sarto2023positive} &	42.9 & 54.8 \\
    EMScore~\cite{shi_emscore_2022}	& 58.9 & 73.9 \\ \midrule
    VC-Inspector-7B & {\bf 72.8} & {\bf 80.3} \\
    \bottomrule
    \end{tabular}
    }
    \vspace{-5pt}
    \caption{Human correlation scores on the YouCook2-Eval dataset.}
    \label{tab:yc2-eval}
\end{table}

\subsection{Correlation with Human Judgments}\label{sec:human_corr}
\paragraph{Evaluation on VATEX-Eval.}
\cref{tab:result_on_vatexeval} presents our primary results: human correlation scores on the VATEX-Eval~\cite{shi_emscore_2022} dataset. 
VATEX-Eval is a widely adopted dataset for evaluating video caption metrics.
It contains 6 captions with varying levels of quality per video, each rated by three human evaluators on a scale of 1 to 5.\footnote{
Since some videos become unavailable, we collect the remaining subset of 2,590 videos and the corresponding 15,540 candidate captions. Unless otherwise specified, all experiments were evaluated on the same dataset to ensure fairness.}  

Baselines are grouped by reference type as outlined in Section~\ref{sec:baselines}, and the three column-sections correspond to the {\it No Reference}, {\it 1-Reference}, and {\it 9-References} settings, respectively. 
Language-based metrics, which rely heavily on textual references, are not applicable to our target, {\it No Reference} setting. Therefore, we focus our comparisons on multimodal methods that incorporate visual inputs.  

\ours consistently outperforms all evaluated metrics in the {\it reference-free} setting, particularly those based on (image-based) CLIP embeddings.
When adapting CLIPScore to the recently introduced ViCLIP~\cite{wang_internvid_2024} (which adopts a video encoder), we observe a considerable gain over those image-CLIP-based approaches.
Despite ViCLIPScore being a stronger baseline, it still underperforms relative to \ours and its underlying model, as the context length of the ViCLIP model (i.e., 32) is much shorter than the 32K context length of ours, which supports a more flexible and comprehensive caption evaluation.
Furthermore, \ours outperforms G-VEval~\citep{tong_g-veval_2024} despite being based on GPT-4o, a substantially larger proprietary model. This demonstrates that explicit factual grounding and explanation supervision are more critical than model scale alone. In contrast, \ours is open-source, lightweight, and reproducible, with configurations available at 3B and 7B parameters depending on system requirements. While having a more compact size relative to G-VEval, our 7B model achieves the highest correlation to human evaluations and could potentially enable on-the-fly quality estimation during training, making it a viable reward model in Reinforcement Learning (RL) applications.
Although we focus on the {\it reference-free} setting, it is noteworthy that \textbf{\ours} even surpasses the performance of most {\it reference-based} metrics.

\begin{table}[t!]
    \centering
    \resizebox{\columnwidth}{!}{
    \setlength{\tabcolsep}{7pt}
    \begin{tabular}{@{}lcc@{}}
        \toprule
         Metric & \textbf{\textit{Flickr8K-Expert}} & \textbf{\textit{Flickr8K-CF}} \\
         \midrule
         \rowcolor{lightgray}\multicolumn{3}{l}{\it Reference-based} \\
         BLEU\_1~\cite{papineni_bleu_2002} & 32.2 & 17.9 \\
         BLEU\_4~\cite{papineni_bleu_2002} & 30.6 & 16.9 \\
         ROUGE~\cite{lin_rouge_2004} &  32.1 & 19.9 \\
         METEOR~\cite{banerjee_meteor_2005} & 41.5 & 22.2 \\
         CIDEr~\cite{vedantam_cider_2015} & 43.6 & 24.6 \\         
         SPICE~\cite{anderson_spice_2016} & 51.7 & 24.4 \\
         BERTScore~\cite{bert-score} & - & 22.8 \\
         CLIPScore~\cite{hessel_clipscore_2022} & 52.6 & 36.4 \\
         PAC-S~\cite{sarto2023positive} & 55.5 & 37.6 \\
         FLEUR~\cite{lee_fleur_2024} & 51.6 & \underline{38.8} \\         
         HICE-S~\cite{zeng_hicescore_2024} & \underline{57.2} & 38.2 \\
         PAC-S++ (ViT-B)~\cite{sarto2025positive} & 55.3 & 37.9 \\
         PAC-S++ (ViT-L)~\cite{sarto2025positive} & - & \underline{38.8} \\
         Polos~\cite{wada_polos_2024} & 56.4 & 37.8 \\
         \rowcolor{lightgray}\multicolumn{3}{l}{\it Reference-free} \\
         CLIPScore~\cite{hessel_clipscore_2022} & 51.1 & 34.4 \\
         PAC-S~\cite{sarto2023positive} & 53.9 & 36.0 \\
         FLEUR~\cite{lee_fleur_2024} & 52.7 & 38.6 \\         
         HICE-S~\cite{zeng_hicescore_2024} & 55.9 & 37.2 \\         
         PAC-S++ (ViT-B)~\cite{sarto2025positive} &  54.1 & 37.0 \\
         PAC-S++ (ViT-L)~\cite{sarto2025positive} & - & 38.5 \\
         \midrule
         {\ours-3B} & {59.9} & 39.0 \\
         {\ours-7B} & \underline{\textbf{63.4}} & \underline{\textbf{46.0}} \\
         \bottomrule
    \end{tabular}
    }
    \vspace{-5pt}
    \caption{Correlation score ($\tau_b$) with human judgments on Flickr8k-Expert and Flickr8k-CF~\cite{hodosh2013framing} dataset. The overall best scores are \textbf{bolded} and best of each section is \underline{underlined}. \textbf{Our model outperforms even the reference-based methods.}}
    \label{tab:flickr_image}
\end{table}

\paragraph{Evaluation on YouCook2-Eval.} 

Due to the limited availability of public benchmarks for video caption metric evaluation, we construct another video caption dataset with human ratings, \texttt{YouCook2-Eval}, based on YouCook2~\cite{ZhXuCoCVPR18} validation set.
We randomly sample 20 videos, each paired with 5 candidate captions including the ground truth caption, a caption from another video, and captions generated from three different models (i.e., Gemini2.5~\cite{comanici2025gemini}, Qwen2.5-VL~\cite{bai_qwen25-vl_2025}, PDVC~\cite{wang2021end}), resulting in 100 video-caption pairs with diverse quality.
Following VATEX-Eval~\cite{shi_emscore_2022}, we collect human ratings from 3 human annotators on a 1-5 Likert scale, and evaluate the correlation to human judgments. 
As shown in \cref{tab:yc2-eval}, \ours achieves the highest correlation compared to baselines, showing generalization to a new domain and caption sources.

\paragraph{Evaluation on Flickr8K.}

We further extend our evaluation to two popular image caption datasets, Flickr8K-Expert and Flickr8K-CF~\cite{hodosh2013framing}, by viewing images as single-frame videos. The former contains 17K image-caption pairs rated by three human experts on a scale of 1 to 4, which requires a more fine-grained differentiation among the captions, whereas the latter collects binary quality assessments for 48K image-caption pairs from crowd sources.
\cref{tab:flickr_image} reports the human correlation score on these two datasets.
\ours is instruction-tuned with captions that exhibit varying degrees of factual inaccuracies, enabling nuanced evaluation of caption quality. This allows the model to perform effectively across both benchmarks, despite their differing rating scheme.
In these evaluations, \ours remains the best-performing method under the {\it reference-free} setting,
and even outperforming several {\it reference-based} metrics. These results demonstrate the strong generalization capability of \ours across visual domains and video lengths.

\subsection{Sensitivity to Hallucinations}\label{sec:foil}
\begin{table}[t!]
    \centering
    \resizebox{\columnwidth}{!}{
    \setlength{\tabcolsep}{8pt}
    \begin{tabular}{@{}lcc@{}}
        \toprule
         Metric & \textbf{\textit{FOIL-COCO}} & \textbf{\textit{ActivityNet-FOIL}} \\
         \midrule
         \rowcolor{lightgray}\multicolumn{3}{l}{\it Reference-based} \\
         EMScore~\cite{shi_emscore_2022} & - & 92.4  \\
         PAC-S~\cite{sarto2023positive} & 93.5 & 93.4 \\
         FLEUR~\cite{lee_fleur_2024} & 97.3 & - \\
         \rowcolor{lightgray}\multicolumn{3}{l}{\it Reference-free} \\
         EMScore~\cite{shi_emscore_2022} & - & 89.5 \\
         PAC-S~\cite{sarto2023positive} & 90.2 & 91.0 \\
         FLEUR~\cite{lee_fleur_2024} & 96.8 & - \\
         \midrule
         {\ours-3B} & {\bf 99.6} & {\bf 99.3} \\
         \bottomrule
    \end{tabular}
    }
    \vspace{-5pt}
    \caption{Accuracy on image/video caption hallucination benchmarks.}
    \label{tab:foil}
\end{table}

\cref{tab:foil} reports accuracy on two hallucination detection benchmarks, FOIL-COCO~\cite{shekhar2017foil} and ActivityNet-FOIL~\cite{shi_emscore_2022}, where the task is to classify FOIL captions from their corresponding correct captions. Each FOIL caption differs from the original by exactly one object error, enabling a controlled evaluation of hallucination sensitivity.
As \ours is explicitly designed to assess object and action grounding in visual-caption alignment, it performs strongly on both image-based (FOIL-COCO) and video-based (ActivityNet-FOIL) benchmarks. These results indicate that \ours is highly sensitive to targeted hallucination errors and can reliably identify factual inconsistencies across varying video lengths and visual contexts.

\subsection{Ablation on Data Synthesis Strategies}\label{sec:ablate_data_syn}

Since both objects and actions are key factual elements in video captions, grounding the evaluator in factual understanding requires instructing the model to identify errors in these components. To this end, we systematically altered both elements in ground truth captions during data generation to create pseudo captions for training.
In this section, we ablate the impact of modifying these elements in the ActivityNet training set by evaluating three variants: i) Changing objects only, ii) Changing actions only, and iii) Changing both. As shown in \cref{tab:ablation_data_syn}, all variants exhibit strong alignment with human ratings compared to EMScore~\cite{shi_emscore_2022}. However, the variant that alters both objects and actions yields the best performance. These results underscore the importance of both factual elements, objects and actions, for capturing video context in caption evaluation. They also demonstrate the robustness and generalization of the proposed paradigm across different factual errors.

\begin{table}[t!]
    \centering
    \resizebox{\columnwidth}{!}{
    \setlength{\tabcolsep}{8pt}
    \begin{tabular}{@{}l l c c@{}}
    \toprule
    Metric & Data Synthesis & $\tau_b$ & $\rho$ \\ 
    \midrule
    EMScore~\cite{shi_emscore_2022}    & n/a           & 22.88 & 29.79 \\
    \midrule
    \multirow{3}{*}{\ours-3B} & Change objects only         & 36.40 & 41.20 \\
     & Change actions only        & 33.23 & 39.63 \\
     & Change both (Ours)   & \textbf{37.99} & \textbf{42.45} \\
    \bottomrule
    \end{tabular}
    }
    \vspace{-5pt}
    \caption{Ablation study on synthetic data generation strategies. Human correlation scores ($\tau_b$, $\rho$) are reported on the VATEX-Eval benchmark.}
    \label{tab:ablation_data_syn}
\end{table}

\subsection{Analysis of Explanations}\label{sec:ablate_exp}

\paragraph{Impact of explanation on performance.} \label{sec:exp_performance}

Our data generation process produces not only pseudo captions and quality scores but also an informative ``side product": explanations identifying factual errors in candidate captions.
We leverage these explanations as auxiliary supervision during training, enabling the model to learn better factual grounding, which has been shown to be effective compared to the variant without explanations in \cref{tab:ablation_exp}. Beyond training benefits, explanations enhance the interpretability of the model-based metric, providing transparency of why specific scores are assigned. %

\paragraph{Using explanations for caption refinement.} \label{sec:exp_vlm_caption}

Having shown that explanations aids in evaluating caption quality, we now ask: {\bf can they also help improve the captions themselves?}
To assess this, we adopt \texttt{Qwen2.5-VL}-7B as the captioning model and use \ours-7B as the critique model that provides feedback during an iterative refinement process, repeated for up to 10 iterations. The prompt used for refinement is provided in \ref{prompt:refine_caption}.
We conduct this experiment on CAPability \citep{liu2025capability}, a recent benchmark that evaluates caption quality across 12 distinct dimensions and two modalities (image and video), with human-annotated ground truth.
Following its protocol, final captions are scored by \texttt{GPT-4.1}.
Note that we report dimensions related to object semantics and events in \cref{tab:capability_f1_results}, and provide the full results in \cref{app:capability}.
We compare against baselines using the same captioner but with different feedback strategies: (a) no feedback, and (b) feedback from the off-the-shelf \texttt{Qwen2.5-VL}-7B.
Results show that the captioner guided by \ours consistently improves over both baselines on these dimensions, demonstrating the effectiveness of explanations for caption refinement. Notably, comparing the performance of (b) to (a), we find that undesired feedback to the captioner can even hurt the performance, significantly. This further underscores the importance of explicit factual grounding in \ours.
\cref{app:capability} additionally presents a qualitative example of the refinement, showing that \ours successfully guides the correction of factual errors in the initial caption.

\begin{table}[t!]
    \centering
    \resizebox{.72\columnwidth}{!}{
    \setlength{\tabcolsep}{9pt}
    \begin{tabular}{@{}l c c@{}}
    \toprule
    \ours-3B & $\tau_b$ & $\rho$ \\
    \midrule
    Without Explanations        & 34.29 & 38.18 \\
    With Explanations           & \textbf{37.99} & \textbf{42.45} \\
    \bottomrule
    \end{tabular}
    }
    \caption{Impact of explanations on the model performance on VATEX-Eval. In the ``Without Explanation" setting, we have trained the model only on the score, removing the pseudo explanations.}\label{tab:ablation_exp}
\end{table}

\paragraph{Evaluation of explanation quality.} \label{sec:exp_human_eval}
For a more direct assessment of explanation quality, we randomly sample 50 captions from the VATEX-Eval dataset, and manually evaluate the explanations generated by \ours-7B.
For each explanation, two human evaluators are provided with the corresponding video and candidate caption, and rate the explanation on a 1-5 Likert scale rather than binary judgments (e.g., good vs. bad), as explanations may be correct in some aspects while missing others. 
To mitigate the potential bias toward assigning high ratings, we introduced a control condition where, for half of evaluation data, the explanation was randomly paired with a different video caption. The scores are normalized to $[0,1]$, and the final score is computed as the score on correct pairs, discounted by the score assigned to incorrect pairs, i.e., $final\textunderscore score = (score~on~correct~pairs) \times (1 - score~on~incorrect~pairs)$. \ours achieves a final score of 0.62, with 86\% inter-evaluator agreement within one point. In addition to human evaluation, we report language model based evaluation in \cref{app:exp_eval}.

\begin{table}[t]
\centering
\small
\resizebox{\columnwidth}{!}{
\setlength{\tabcolsep}{5pt}
\begin{tabular}{@{}lccc@{}}
\toprule
& \multicolumn{3}{c}{Feedback Provider} \\
\cmidrule{2-4}
\textbf{Task} & None & \texttt{Qwen2.5-VL}-7B & \ours-7B \\
\midrule
\rowcolor{blue!10}
\rowcolor{blue!10}
Dynamic Object Num.   & 17.4 & 17.3 & \textbf{20.7} \\
\rowcolor{blue!10}
Event              & 58.2 & 18.8 & \textbf{58.6} \\
\rowcolor{orange!10}
Object Category     & 67.8 & 58.9 & \textbf{68.0} \\
\rowcolor{orange!10}
Object Number       & 28.8 & 20.4 & \textbf{29.3} \\
\rowcolor{orange!10}
Spatial Relation    & 56.6 & 45.7 & \textbf{57.1} \\
\bottomrule
\end{tabular}
}
\vspace{-4pt}
\caption{F1 scores across different tasks. Video-based tasks are highlighted in \textcolor{blue!80}{blue}, while image-based tasks are highlighted in \textcolor{orange!80}{beige}. Best results are shown in \textbf{bold}.}
\label{tab:capability_f1_results}
\end{table}

\subsection{Computational Efficiency}
We assess computational efficiency by comparing the average runtime per video. On a single A100 GPU, EMScore \citep{shi_emscore_2022}, ViCLIPScore \citep{wang_internvid_2024}, and \ours-3B require 0.42, 0.34, and 0.30 seconds per clip, respectively. These results indicate that \ours is more efficient than existing methods. For EMScore and ViCLIPScore, we use the official implementations provided by the authors; for \ours, we employ vLLM \citep{kwon2023efficient} with online serving. Efficiency is measured over 2,000 videos, each uniformly sampled into 16 frames.

\section{Conclusion} \label{sec:discussion}
This work tackles the challenge of evaluating video captions across diverse domains without relying on human-annotated references. We conducted an extensive review of existing metrics, identified their limitations, and proposed \ours, a novel {\it fact-aware} evaluator for {\it reference-free} video caption evaluation. By incorporating factual analysis into caption assessment, we equipped a lightweight, open-source LMM with the factual grounding capabilities through training on captions of diverse quality, where pseudo captions with controllable factual errors were systematically generated.
Experimental results across multiple domains demonstrate that \ours achieves high alignment with human judgments, and outperforms existing metrics in detecting factual errors.
Its versatility and interpretability make it a practical tool for assessing factual accuracy in real-world video captioning scenarios.
Furthermore, ablation studies on predicted explanations underscore its effectiveness in quality estimation and caption refinement.

\section*{Limitations}

\ours primarily targets object and action correctness, which constitute major sources of factual errors in video captions. While these factors cover many common failure cases, other important aspects, e.g., attributes, spatial relationships, and fine-grained temporal ordering, are not explicitly modeled and remain directions for future work. Additionally, the training process partially relies on synthetically generated captions and pseudo-scores. Although these scores are deterministic and validated through correlation with human judgments, they may not fully capture the diversity of real-world captioning errors. Finally, while \ours is substantially more lightweight than proprietary LLM-based evaluators, it still depends on a multimodal backbone, resulting in higher computational costs compared to purely embedding-based metrics.

\textit{Despite these limitations, \ours represents a meaningful step toward \textbf{explanation-aware}, \textbf{factually grounded}, and \textbf{reference-free} video caption evaluation, providing a flexible foundation for future extensions.}

\section*{Acknowledgment}
Some experiments were conducted on the UMBC HPCF, supported by the National Science Foundation under Grant No. CNS-1920079. %

\bibliography{normalized}

\appendix

\section*{Appendix}

\section{Qualitative Results}\label{app:visual}

\begin{figure}[t!]
    \centering
    \includegraphics[width=\linewidth]{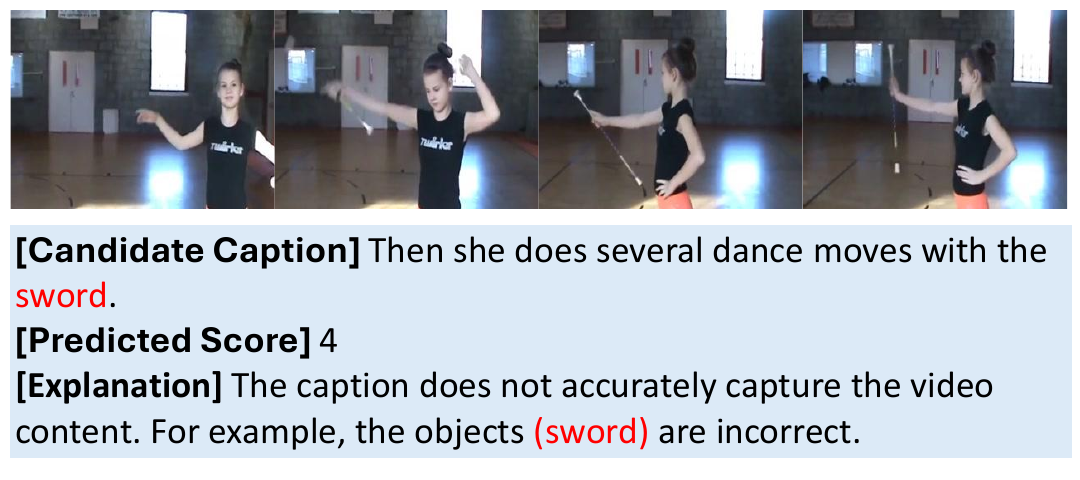} \\
    \vspace{-3pt}
    \includegraphics[width=\linewidth]{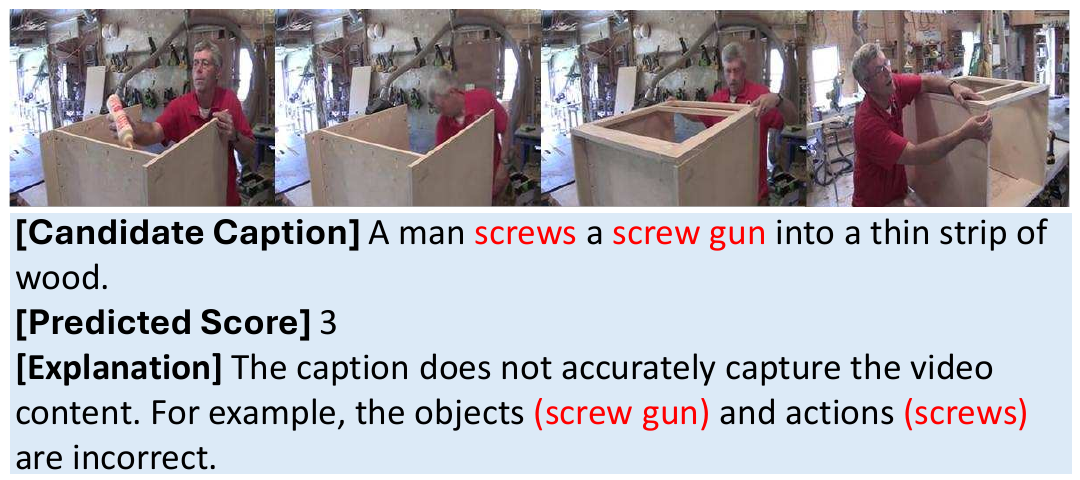} \\
    \includegraphics[width=\linewidth]{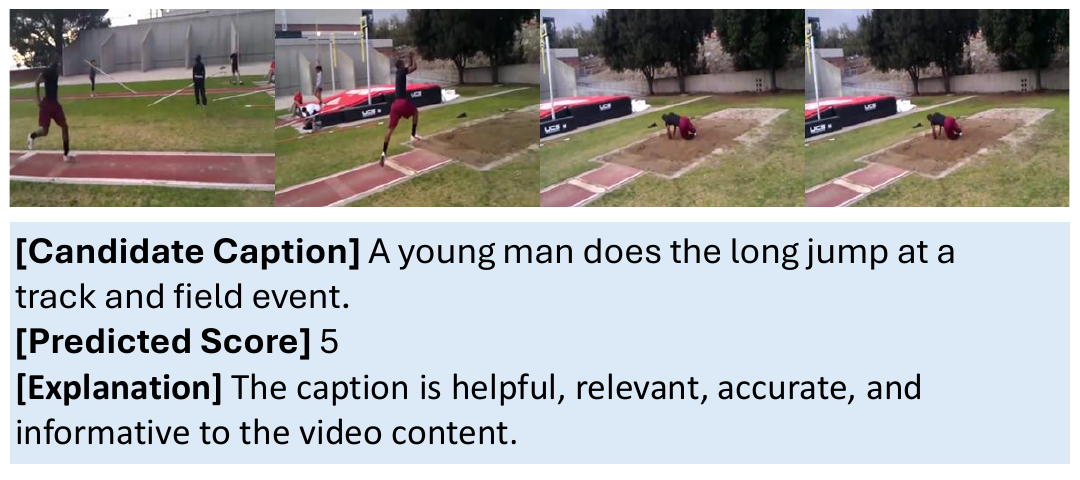} \\
    \captionof{figure}{
    Visual examples 
    from {\tt ActivityNet-FG-Eval} (top) and VATEX-Eval (others). \ours produces quality assessments 
    consistent with ground truth scores, 
    and 
    explanatory insights into factual errors (highlighted in red).}
    \label{fig:qual_example}
\end{figure}

\cref{fig:qual_example} presents visual examples of \ours outputs on the {\tt ActivityNet-FG-Eval} (top) and VATEX-Eval (others) datasets.
Without relying on reference captions, the model evaluates candidate captions based on the associated video content and produces quality scores that closely mirror human judgments.
The explanation effectively pinpoints the factual inaccuracies, such as incorrect objects and/or actions in the candidate captions, and assigns scores accordingly.

\begin{figure}[ht!]
    \centering
    \label{fig:additional_vatex_example}
    \begin{minipage}[h]{\linewidth}
        \includegraphics[width=\linewidth]{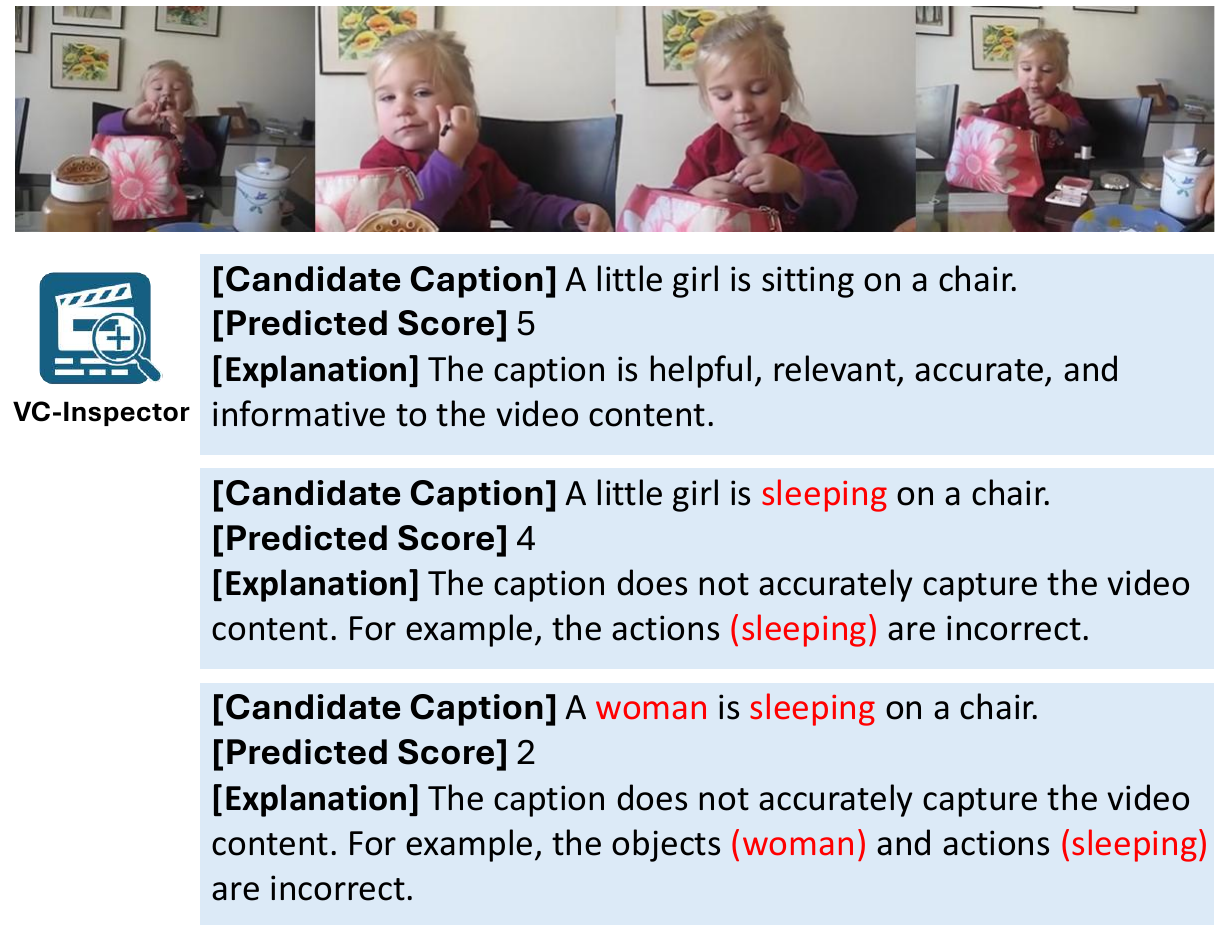}
    \end{minipage}
    \hfill
    \begin{minipage}[h]{\linewidth}
        \includegraphics[width=\linewidth]{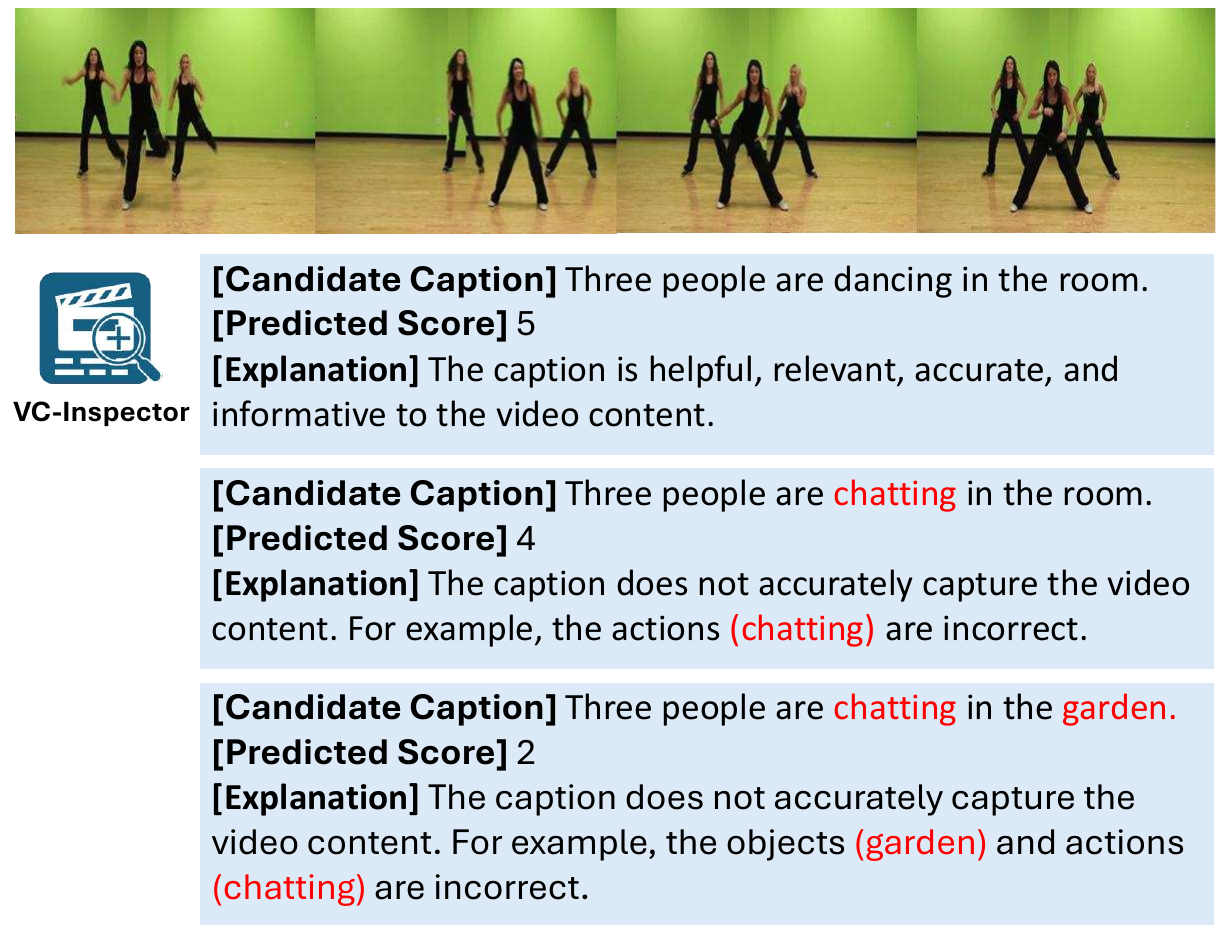}
    \end{minipage}  
    \captionof{figure}{Additional visualization of \ours results on {\tt ActivityNet-FG-Eval} videos, with candidate captions of diverse quality. Incorrect objects and actions are identified by \ours and labeled in red.}
    \label{fig:additional_anet_example}
\end{figure}

In \cref{fig:additional_anet_example}, we showcase two video examples from {\tt ActivityNet-FG-Eval}, each paired with candidate captions containing a progressively increasing number of factual errors. The model successfully detects these incorrect elements, provides detailed explanations, and yields scores that reflect the severity of the factual inaccuracies.

\section{Additional Results}\label{app:ablation}

\subsection{Evaluation of Explanation Quality using Language Models}\label{app:exp_eval}

We further assess the quality of the explanations generated by \ours on two synthetic datasets, as the VATEX-Eval dataset does not offer ground truth explanations. Table~\ref{tab:exp_eval} presents two metrics for this experiment. The BERT score measures the semantic similarity between the generated explanations and the pseudo ground truth explanations, according to pretrained textual embeddings. On the other hand, the LLM Score is derived by using the prompt detailed in Section~\ref{app:prompts} to engage {\tt Llama-3-8B-Instruct} in a LLM-as-a-judge approach, similar to \cite{liu2023llava}. 
We report the relative score assigned by the LLM that rescales the score received by the predicted explanations with respect to the score given to the ground truth explanations.
Results show that the explanations provided by \ours, along with its quality assessment, align closely with the (pseudo) ground truth explanations obtained during the data generation process.

\subsection{Complete Results for Caption Refinement}\label{app:capability}
In Section~\ref{sec:exp_vlm_caption}, we present the experiment of caption refinement with CAPability benchmark, where we report the dimensions concerning objects and actions, as they are the primary focus of this work.
We provide the F1 scores of all the 12 dimensions in 
\cref{tab:capability_f1_results_full} for completeness.  
In addition to quantitative results, \cref{fig:qual-dynobj-80} visualizes an example of the refinement, showing the initial and refined captions produced under guidance from \ours.

\begin{table}[t!]
    \centering
    \resizebox{.8\columnwidth}{!}{
    \setlength{\tabcolsep}{9pt}
    \begin{tabular}{@{}l|cc@{}}
        \toprule
         Dataset &  BERT Score & LLM Score \\ \midrule
         ActivityNet-FG-Eval & 0.79 & 93.11 \\
         YouCook2-FG-Eval & 0.70 & 90.97 \\
         \bottomrule
    \end{tabular}
    }
    \caption{Evaluation of the explanations generated by \ours-3B on two synthetic evaluation datasets.}
    \label{tab:exp_eval}
\end{table}

\begin{table}[t]
\centering
\resizebox{\columnwidth}{!}{
    \begin{tabular}{@{}lccc@{}}
    \toprule
    & \multicolumn{3}{c}{Feedback Provider} \\
    \cmidrule{2-4}
    \textbf{Task} & None & \texttt{Qwen2.5-VL}-7B & \ours-7B \\
    \midrule
    \rowcolor{blue!10}
    Camera Angle        & \textbf{37.5} & 6.4  & 36.1 \\
    \rowcolor{blue!10}
    Camera Movement     & 26.4 & 0.6  & \textbf{29.3} \\
    \rowcolor{orange!10}
    Character Identification    & 29.6 & 17.3 & \textbf{30.5} \\
    \rowcolor{blue!10}
    Dynamic Object Num.   & 17.4 & 17.3 & \textbf{20.7} \\
    \rowcolor{blue!10}
    Event               & 58.2 & 18.8 & \textbf{58.6} \\
    \rowcolor{orange!10}
    Object Category     & 67.8 & 58.9 & \textbf{68.0} \\
    \rowcolor{orange!10}
    Object Color        & \textbf{73.0} & 51.9 & 72.3 \\
    \rowcolor{orange!10}
    Object Number       & 28.8 & 20.4 & \textbf{29.3} \\
    \rowcolor{orange!10}
    OCR                 & \textbf{69.6} & 62.9 & 68.5 \\
    \rowcolor{orange!10}
    Scene               & \textbf{63.5} & 63.0 & 63.2 \\
    \rowcolor{orange!10}
    Spatial Relation    & 56.6 & 45.7 & \textbf{57.1} \\
    \rowcolor{orange!10}
    Style               & \textbf{84.3} & 78.7 & 84.2 \\
    \midrule
    \textbf{Average}    & 51.1 & 36.8 & \textbf{51.5} \\
    \bottomrule
    \end{tabular}
    }
\caption{F1 scores across different tasks. Video-based tasks are highlighted in \textcolor{blue!80}{blue}, while image-based tasks are highlighted in \textcolor{orange!80}{beige}. Best results are shown in \textbf{bold}.}
\label{tab:capability_f1_results_full}
\end{table}

\section{Training Hyperparameters} \label{app:train_hyperparams}
We report all training hyperparameters in \cref{tab:hyperparameters}.

\begin{table}[t]
\centering
\small
\begin{tabular}{@{}lr@{}}
\toprule
\textbf{Hyperparameter} & \textbf{Value} \\
\midrule
\multicolumn{2}{@{}l@{}}{\textit{Training Configuration}} \\
Epochs & 1 \\
Global Batch Size & 128 \\
Learning Rate & 1e-4 \\
LR Scheduler & Cosine (min: 1e-5) \\
Warmup Ratio & 0.05 \\
\midrule
\multicolumn{2}{@{}l@{}}{\textit{LoRA Configuration}} \\
Rank ($r$) & 32 \\
Alpha ($\alpha$) & 32 \\
Dropout & 0.05 \\
\midrule
\multicolumn{2}{@{}l@{}}{\textit{Module Frozen}} \\
Vision Encoder & \ding{51} \\
Language Model & \ding{55} \\
Projector & \ding{51} \\
\midrule
\multicolumn{2}{@{}l@{}}{\textit{Video Processing}} \\
Number of Frames & 32 \\
Max. Pixels & $224 \times 224$ \\
\bottomrule
\end{tabular}
\caption{Hyperparameters for instruction fine-tuning.}
\label{tab:hyperparameters}
\end{table}

\section{Prompts} \label{app:prompts}
The data generation prompts are reported on the following blocks:
\begin{itemize}[itemsep=0pt]
    \item Extract object -- Prompt~\ref{prompt:obj_list}
    \item Extract action -- Prompt~\ref{prompt:act_list}
    \item Find similar object -- Prompt~\ref{prompt:new_obj}
    \item Find similar action -- Prompt~\ref{prompt:new_act}
    \item Substitute object or action -- Prompt~\ref{prompt:substitute}
\end{itemize}

The data generation process provides the information of incorrect objects and actions.
We format the explanations of these factual errors using the following template:
\begin{itemize}[itemsep=0pt]
    \item Captions without errors: The caption is helpful, relevant, accurate, and informative to the video content.
    \item Captions with errors: The caption does not accurately capture the video content. For example, the actions (\{\{wrong\textunderscore act\}\}) are incorrect / the objects (\{\{wrong\textunderscore obj\}\}) are incorrect / the objects (\{\{wrong\textunderscore obj\}\}) and actions (\{\{wrong\textunderscore act\}\}) are incorrect.
\end{itemize}
These explanations are then used in Prompt~\ref{prompt:fine_tuning_prompt} for training \ours.

The prompt for evaluating the generated explanations is in Prompt~\ref{prompt:exp_prompt}. Prompt~\ref{prompt:refine_caption} is used to refine the caption based on the explanation.

\begin{promptbox}[prompt:obj_list]{Extract object from a caption}
\begin{Verbatim}[breaklines=true,breaksymbol=,fontfamily=tt]
### Instruction:

Given the input text, generate a list of objects in the caption in the format of [ "Object1", "Object2", ...]. Don't include any verbs. ONLY REPLY THE ANSWER.

### Input: {{caption}}
### Output:
\end{Verbatim}
\end{promptbox}

\begin{promptbox}[prompt:act_list]{Extract actions from a caption}
\begin{Verbatim}[breaklines=true,breaksymbol=,fontfamily=tt]
### Instruction:

Given the input text, generate a list of actions in the caption in the format of ["Action1", "Action2", ...]. ONLY REPLY THE ANSWER.

### Input: {{caption}}

### Output:
\end{Verbatim}
\end{promptbox}

\begin{promptbox}[prompt:new_obj]{Find similar object given an object}
\begin{Verbatim}[breaklines=true,breaksymbol=,fontfamily=tt]
### Instruction:

Find the parent class of the given object and generate one of its child classes that has a different meaning but shares the same parent. The new class cannot be a synonym or similar term to the original object. It can be an antonym or any co-hyponym. For example, generate "dog" for "cat". ONLY REPLY THE NEW CLASS.

### Input: {{object}}

### Output:
\end{Verbatim}
\end{promptbox}

\begin{promptbox}[prompt:new_act]{Prompt to find similar action given an action}
\begin{Verbatim}[breaklines=true,breaksymbol=,fontfamily=tt]
### Instruction:

Find a different action that the subject can perform that has a different meaning than the input action. The new action cannot be a synonym or similar term to the original action. For example, generate "put into" for "take out of". ONLY REPLY THE NEW ACTION.

### Input: {{action}}

### Output:
\end{Verbatim}
\end{promptbox}

\begin{promptbox}[prompt:substitute]{Prompt to substitute object or action given the caption and new object or actions}
\begin{Verbatim}[breaklines=true,breaksymbol=,fontfamily=tt]
### Instruction:

Substitute {{old_obj_act}} in {{caption}} as {{new_obj_act}}. Keep the answer in the same format as {{caption}}. ONLY REPLY THE ANSWER.

### Input: {{caption}}

### Output:
\end{Verbatim}
\end{promptbox}

\begin{promptbox}[prompt:fine_tuning_prompt]{Fine-tuning prompt}
\begin{Verbatim}[breaklines=true,breaksymbol=,fontfamily=tt]
### USER:
{{VIDEO}}

<caption>{{caption}}</caption> You are given a video and a caption describing the video content. Please rate the helpfulness, relevance, accuracy, level of detail of the caption. The overall score should be on a scale of 1 to 5, where a higher score indicates better overall performance. Please first output a single line containing only one integer indicating the score. In the subsequent line, please provide a comprehensive explanation of your evaluation, avoiding any potential bias. STRICTLY FOLLOW THE FORMAT.

### ASSISTANT:
{{quality_score}}

{{explanation}}
\end{Verbatim}
\end{promptbox}

\begin{promptbox}[prompt:exp_prompt]{Prompt for explanation evaluation}
\begin{Verbatim}[breaklines=true,breaksymbol=,fontfamily=tt]
[Context]
{{ground_truth_caption}}

[Caption]
{{caption_to_evaluate}}

[Groundtruth]
{{ground_truth_explanation}}
[End of Groundtruth]

[Assistant]
{{predicted_explanation}}
[End of Assistant]

[System]
We would like to request your feedback on the performance of an AI assistant in the response to the quality evaluation of the caption provided above with respect to a video. For your reference, the visual content in the video is represented with a few sentences describing the same video. You are also given a ground truth evaluation to that caption.
Please rate the helpfulness, relevance, accuracy, level of detail of the response by comparing to the ground truth and referring to the context information. Provide an overall score on a scale of 1 to 10, where a higher score indicates better overall performance.  

Please first output a single line containing the score. In the subsequent line, please provide a comprehensive explanation of your evaluation, avoiding any potential bias.
\end{Verbatim}
\end{promptbox}

\begin{promptbox}[prompt:refine_caption]{Prompt to refine caption based on explanation}
\begin{Verbatim}[breaklines=true,breaksymbol=,fontfamily=tt]
Previous Caption: {{caption}}
Evaluation: {{explanation}}

Based on the previous caption and evaluation, refine the caption to make it more accurate and detailed.
Use the evaluation to find out the missing details and use the video frames to fix the missing details.
Just output the refined caption, without any other text.
\end{Verbatim}
\end{promptbox}

\section{Use of AI Assistance}
During the preparation of this work, the authors utilized Cursor\footnote{\url{https://cursor.com}} for coding assistance and ChatGPT\footnote{\url{https://chatgpt.com/}} for proofreading and language refinement. All input provided to these tools consisted of original content authored by the authors.

\begin{figure*}[t]
\centering
\footnotesize
\renewcommand{\arraystretch}{1.3}
\setlength{\tabcolsep}{4pt}
\begin{tabular}{@{} >{\centering\arraybackslash}m{0.085\linewidth} p{0.885\linewidth} @{}}
\toprule
\multicolumn{2}{c}{%
    \begin{tabular}{cccccccc}
      \includegraphics[width=0.100\linewidth,height=0.155\linewidth,keepaspectratio]{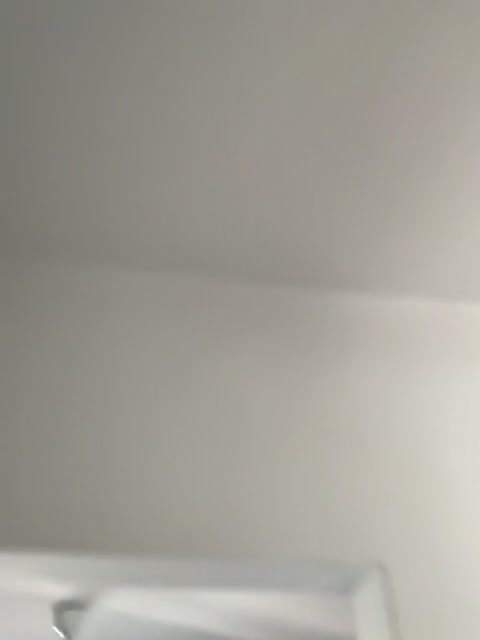} &
      \includegraphics[width=0.100\linewidth,height=0.155\linewidth,keepaspectratio]{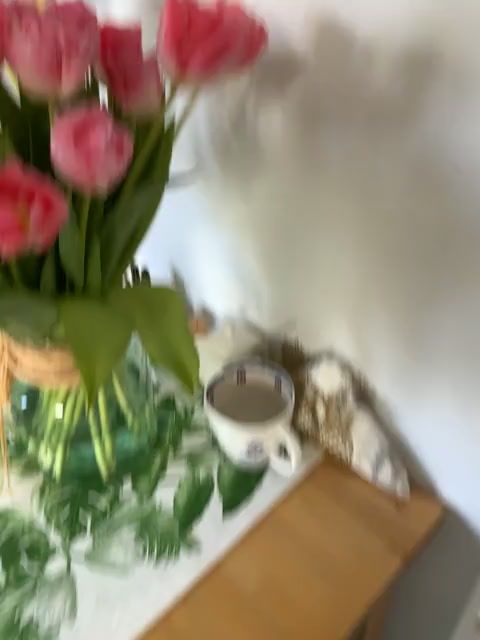} &
      \includegraphics[width=0.100\linewidth,height=0.155\linewidth,keepaspectratio]{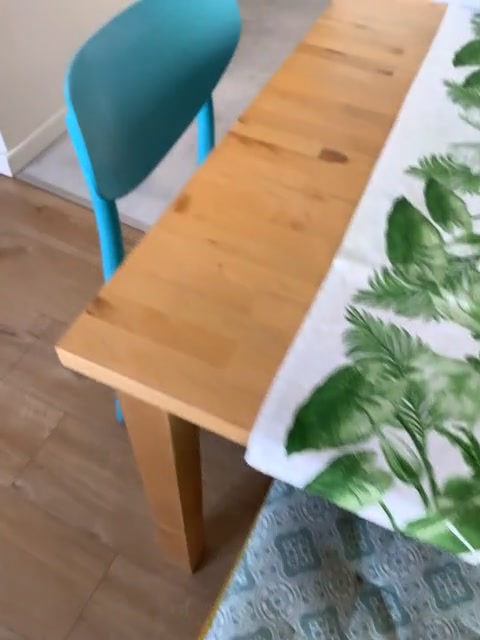} &
      \includegraphics[width=0.100\linewidth,height=0.155\linewidth,keepaspectratio]{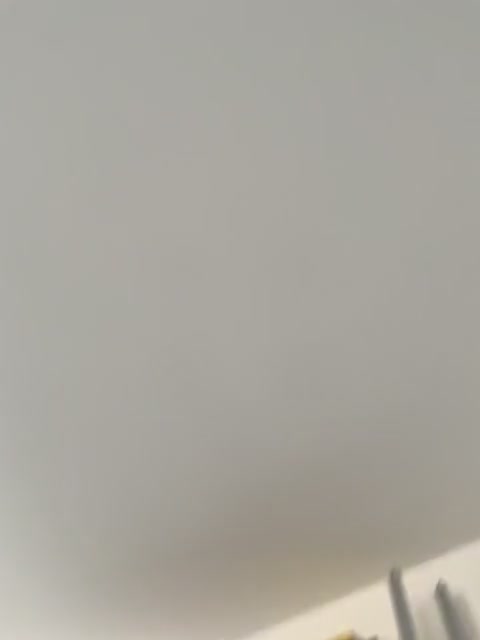} &
      \includegraphics[width=0.100\linewidth,height=0.155\linewidth,keepaspectratio]{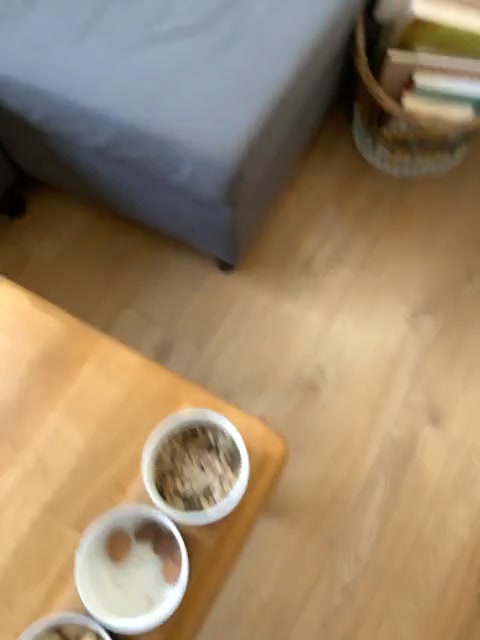} &
      \includegraphics[width=0.100\linewidth,height=0.155\linewidth,keepaspectratio]{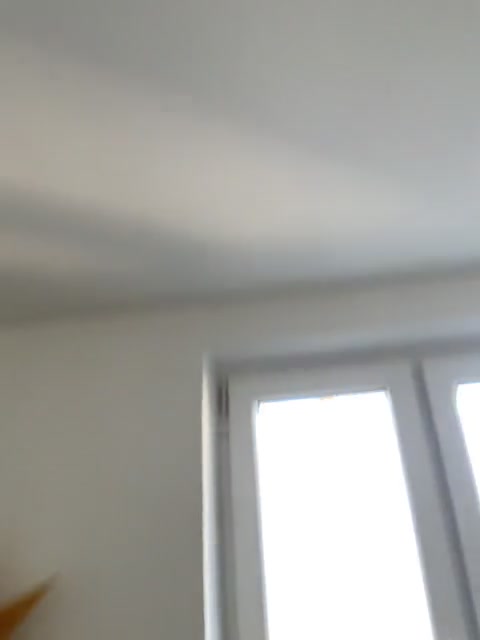} &
      \includegraphics[width=0.100\linewidth,height=0.155\linewidth,keepaspectratio]{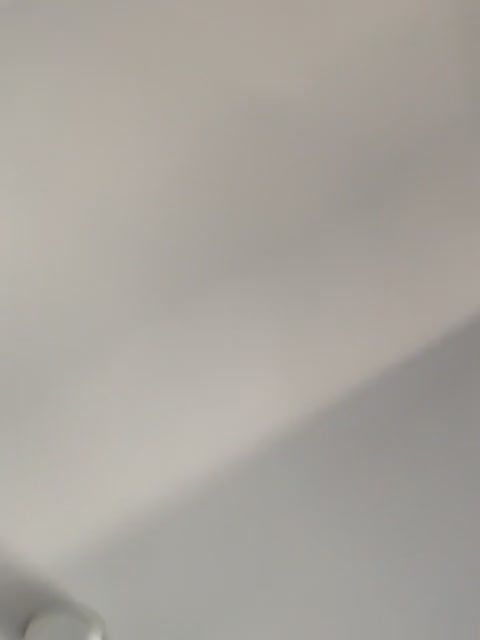} &
      \includegraphics[width=0.100\linewidth,height=0.155\linewidth,keepaspectratio]{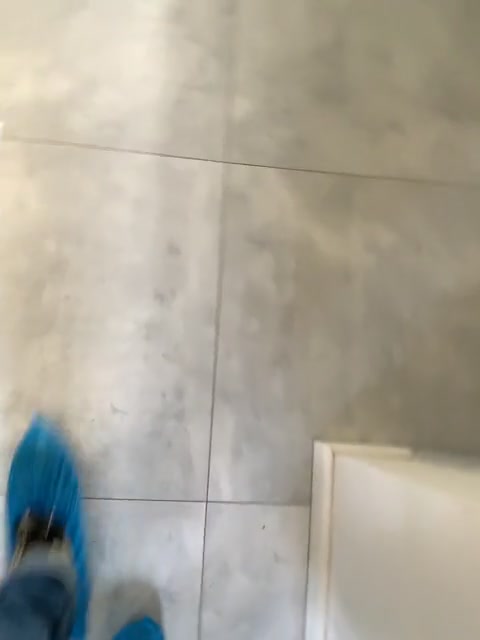} \\
      $t{=}2.7$\,s & $t{=}8.2$\,s & $t{=}13.6$\,s & $t{=}19.1$\,s & $t{=}24.5$\,s & $t{=}30.0$\,s & $t{=}35.5$\,s & $t{=}40.9$\,s \\
    \end{tabular}} \\
\midrule
\makecell{\textbf{Round 0}\\[1pt]\textcolor{acllow}{\textbf{3\,/\,5}}} &
The video begins with a clear, focused shot of a vintage-style poster for the ``Chicago World's Fair'' from 1933-1934, which is laid flat on a light gray tiled floor. The poster features a striking red background with an illustration of the fair's iconic buildings and the text ``CHICAGO WORLD'S FAIR MAY 27 - NOV. 1, 1933'' at the top.

The camera then moves erratically, capturing blurry glimpses of the surrounding environment. It pans across a hallway or room corner, showing a white wall and a glimpse of a wooden door frame. The view shifts to a dining area where a wooden table is covered with a green patterned tablecloth, accompanied by a blue chair. A vase with pink flowers sits on the table, and a potted plant is visible nearby.

The camera continues its shaky movement, briefly focusing on \errhl{a yellow chair with a patterned seat cushion}, before moving to show a dark-colored sofa with pillows against a wall. \errhl{A small side table with books is next to the sofa.} The camera then pans upward, showing the ceiling and a light fixture.

The perspective changes again, revealing a yellow ottoman near a wooden cabinet or shelving unit. The camera then moves back toward the hallway, showing the tiled floor and a glimpse of a doorway leading to another room. The final frames return to the poster, now viewed from a slightly different angle, still lying on the tiled floor, with a person's blue shoe visible in the foreground. The video ends with a clear, steady shot of the poster, emphasizing its design and historical significance. \\
\addlinespace[2pt]
\midrule
\makecell{\textbf{Round 5}\\[1pt]\textcolor{aclhigh}{\textbf{5\,/\,5}}} &
The video opens with a clear, steady shot of a vintage-style poster for the ``Chicago World's Fair'' from 1933-1934, laid flat on a light gray tiled floor. The poster features a striking red background with an illustration of the fair's iconic buildings and the text ``CHICAGO WORLD'S FAIR MAY 27 - NOV. 1, 1933'' at the top.

The camera then moves erratically, panning and tilting rapidly through various parts of a modern, well-lit apartment interior. Due to the quick, unsteady movements, many shots are blurry, but glimpses of different areas and objects can be discerned:

\begin{itemize}[leftmargin=1.1em,itemsep=1pt,topsep=2pt,parsep=0pt]
  \item A hallway with a white wall and a wooden door frame.
  \item A dining area featuring a wooden table, a bright blue chair, and a vase of pink flowers.
  \item A living room with a dark gray sofa, \fixhl{a yellow ottoman}, and \fixhl{a wooden shelf mounted on the wall holding books and decorative items}.
  \item \fixhl{A view of a kitchen area with white cabinets and a glimpse of a refrigerator.}
  \item A close-up of a person's blue shoe on the floor next to the poster.
  \item The poster is shown again, this time from a slightly different angle, still lying on the tiled floor.
\end{itemize}

The video concludes with a final clear shot of the Chicago World's Fair poster, emphasizing its design and historical significance against the backdrop of the contemporary interior. The overall impression is that of a casual, perhaps hurried, tour of a stylishly furnished home, with the poster serving as a point of interest or a decorative element. \\
\bottomrule
\end{tabular}
\caption{Qualitative example from the \textsc{Dynamic Object Number} split of
CAPability captioned by \texttt{Qwen2.5-VL}-7B.
\textbf{Round~0} is the raw caption with no self-refinement.
After five rounds of iterative refinement, \textbf{Round~5} restructures the
shaky mid-section into an explicit room-by-room inventory, and the CAPability
score improves from \textcolor{acllow}{3/5} to \textcolor{aclhigh}{5/5}.
\errhl{Red shading} marks errors in Round~0 that Round~5 corrects:
the yellow ottoman is misidentified as a chair, and the wicker basket of
books is described as a side table.
\fixhl{Green shading} marks the corresponding Round~5 fixes, together with
the newly-covered kitchen area that Round~0 omitted.}
\label{fig:qual-dynobj-80}
\end{figure*}

\end{document}